\documentclass[runningheads]{llncs}
\usepackage[T1]{fontenc}
\usepackage{graphicx}
\usepackage{enumitem}
\usepackage{eso-pic}
\usepackage{amsmath}
\usepackage{cite}
\usepackage{hyperref}  
\begin{document}
\title{BanglaMemeEvidence: A Multimodal Benchmark Dataset for Explanatory Evidence Detection in Bengali Memes}

\titlerunning{BanglaMemeEvidence: A Multimodal Benchmark Dataset}
%
\author{Fatema Tuj Johora Faria\inst{1} \and
Mukaffi Bin Moin\inst{1} \and
Md. Mahfuzur Rahman\inst{1}\and
Pronay Debnath\inst{1}\and
Asif Iftekher Fahim \inst{1} \and
Faisal Muhammad Shah\inst{1}
}
\authorrunning{F.T.J. Faria et al.}
%
\institute{Ahsanullah University of Science and Technology, Dhaka-1208, Bangladesh
\email{fatema.faria142@gmail.com, mukaffi28@gmail.com, pronaydebnath99@gmail.com, mahim1066@gmail.com, fahimthescientist@gmail.com, faisal.cse@aust.edu}
}
\maketitle              
%
\AddToShipoutPictureBG*{%
  \AtPageUpperLeft{%
    \hspace{0.5cm}\raisebox{-1.5cm}{%
      \parbox{\dimexpr\textwidth+8.5cm\relax}{\centering\small
        \textit{\textbf{Accepted at 6th International Conference on Innovations in Computational Intelligence and Computer Vision (ICICV 2026).}}
      }%
    }%
  }%
}
\begin{abstract}
Memes have become influential communication tools on social media, combining viral visuals with concise messaging to convey impactful ideas. While substantial research has examined the affective dimensions of memes, key challenges such as detecting harmful content, identifying cyberbullying, and performing accurate sentiment analysis remain critical, largely due to the need for deeper contextual understanding. In this paper, we introduce \textbf{MemeEvidenceDetect}, a hybrid task aimed at analyzing a meme and its contextual information to identify specific sentences that explain or elucidate its meaning and humor. To support this task, we present \textbf{BanglaMemeEvidence}, a curated dataset of 2,917 Bengali memes, emphasizing its significance as a resource for the Bangla language. Each meme is annotated with natural language explanations, including Meme OCR, Meme Context, and Evidence Sentences, alongside relevance scores that reflect the relationship between a meme and its corresponding annotations. To address the gap in dynamically inferring a meme’s context, we propose \textbf{BengaliMemeEvidenceNet}, a hybrid multimodal framework that integrates textual and visual features for comprehensive meme representation. Our experiments demonstrate the effectiveness of BengaliMemeEvidenceNet, achieving an F1 score of 0.74. To the best of our knowledge, this is the first study to focus on evidence detection in Bengali memes, marking a notable step forward in the analysis of memes in low-resource languages.

\end{abstract}

\section{Intoduction}
Social media has become a key way for people to communicate, changing how we interact in society. The content shared on social media comes in many forms, including text, audio, and images, or a mix of them. Memes, often shared with humor or sarcasm, are a common example. While memes help spread complex ideas about society, culture, or politics, they often lack the context needed for full understanding, which is important for both humans and computers to interpret them accurately. \cite{intro1} \cite{intro2}. 
\begin{figure}[h]
  \includegraphics[width=\columnwidth]{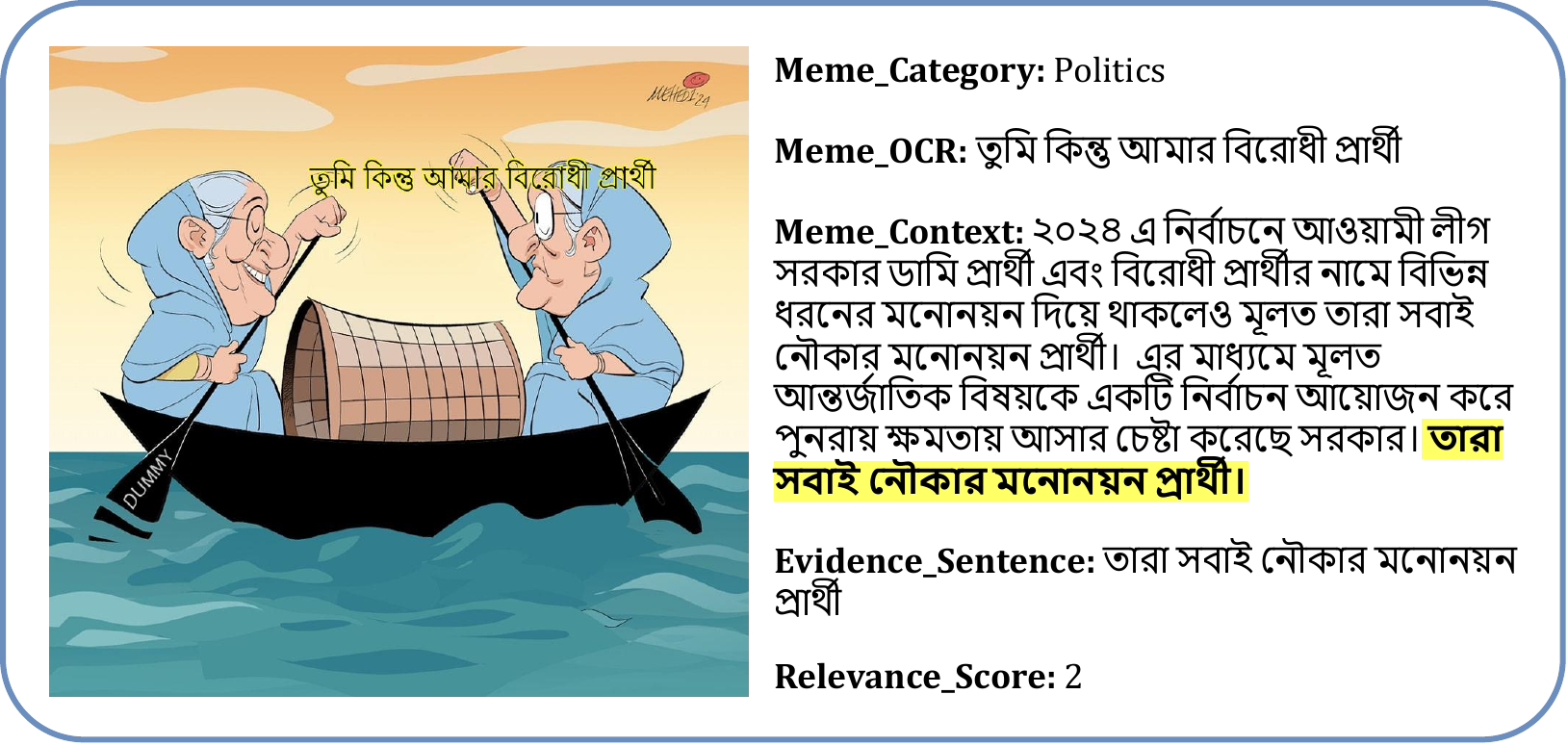}
  \caption{An illustration of MemeEvidenceDetect, a hybrid task where, given a meme, manually transcribed Bengali text, and a description of the meme's context, the meme is supported by evidence sentences. These sentences are tagged with relevance scores, where a score of 2 denotes ``Relevant''.}
  \label{expe}
\end{figure}

The recent surge in meme dissemination has led to a growing body of research on meme analysis, focusing on tasks such as multimodal hate speech detection from Bengali memes and texts \cite{intro3}, the development of datasets for Bengali abusive meme classification \cite{intro4}, multimodal cyberbullying meme detection from social media using deep learning approaches \cite{intro5}, and multimodal analysis of memes for sentiment extraction \cite{intro6}. Other studies have worked on multimodal offensive meme classification using transformers \cite{intro7}, sarcasm detection in typographic memes \cite{intro8}, and multi-task learning frameworks for multi-modal sarcasm, sentiment, and emotion analysis \cite{intro9}. 

Despite the extensive work done in this field, there has been a noticeable lack of emphasis on multimodal evidence detection, especially concerning Bangla memes, which poses a significant challenge. To address this gap, we present innovative solutions aimed at automating the extraction of contextual evidence, thereby enhancing meme accessibility. Introducing a pioneering framework called BengaliMemeEvidenceNet, our approach focuses on identifying sentences within the provided context that can potentially elucidate the meme's meaning.

Figure \ref{expe} presents a meme categorized under Politics, featuring Bengali text that translates to ``\textbf{But you are my opposing candidate.}'' This meme contextualizes the 2024 elections, depicting a complex political scenario in which the Awami League leadership strategically nominates candidates from both its own ranks and rival parties. This strategic move seeks to influence world opinion and strengthen domestic authority by creating the perception of democratic choice while concealing underlying political strategies.
The goal of this research is to analyze this multimodal meme in order to uncover and investigate evidence of political interference during electoral processes. The embedded evidence sentence in the meme provides direct insight into the government's strategy: intentional candidate nominations across the political spectrum to shape global perceptions and consolidate local power. This study seeks to explore methodologies for identifying such evidence within related contextual documents. By doing so, it aims to elucidate the complex dynamics of communication in modern environments across diverse subjects and themes depicted in memes.
To summarize, our main contributions are as follows:
\begin{itemize}[label=$\circ$]

\item \textbf{MemeEvidenceDetect} is an hybrid task that predicts and analyzes the relevance of sentences to a given meme, aiming to clarify its meaning and humor. This task involves detecting relevance scores, with 0 indicating ``Not relevant,'' 1 representing ``Partially relevant,'' and 2 signifying ``Relevant.''

    \item To bridge the gap in Bangla meme evidence research, we introduce a novel dataset called \textbf{BanglaMemeEvidence}, designed to facilitate the automatic detection of contextual evidence in such memes. The dataset comprises gold standard human annotations for 2,917 Bengali memes. The Bengali text from the meme image is manually transcribed to ensure accurate content representation. 
    \item We introduce \textbf{BengaliMemeEvidenceNet}, a  hybrid multimodal framework aimed at identifying evidence for memes from their related contexts. To achieve this, we employ various fusion techniques to integrate text and image features. 
\end{itemize}

\section{Related Work}
\subsection{Meme Analysis} The study by Karim et al. \cite{intro3} constructed a dataset integrating textual and visual elements to tackle hate speech detection. Their study showcased the efficacy of Conv-LSTM for textual analysis and DenseNet-161 for visual interpretation, alongside highlighting the effectiveness of XLM-RoBERTa in textual analysis. In a parallel study \cite{intro4}, delved into abusive memes with the BanglaAbuseMeme dataset, favoring multimodal strategies, particularly CLIP. While XLMR showed promise in textual analysis and ViT in visual, it was CLIP that achieved the highest macro score. Transitioning to cyberbullying detection, Ahmed et al. \cite{intro5} proposed a VGG16-BiLSTM model, achieving notable accuracy in detecting harmful content within Bengali memes. Meanwhile, Alluri and Krishna \cite{intro6} explored meme classification using the Memotion dataset, employing ViT for visuals and RoBERTa for textual analysis, alongside a transformer-based image captioning model for contextual understanding. 

\subsection{Visual Question Answering (VQA)}
Recent advancements in Bengali Visual Question Generation (VQG) and Visual Question Answering (VQA) underscore the need for further exploration in these areas. Hasan et al. \cite{hasan2023visual} have made significant strides in Bengali VQG by employing a transformer-based approach and introducing novel models, such as image-category and image-answer-category variants, which have demonstrated superior question generation capabilities. In contrast, Islam et al. \cite{islam2022note} focused on Bengali VQA by adapting existing datasets and employing a hybrid model that combines CNN and bi-LSTM layers. Meanwhile, Rafi et al. \cite{rafi2022deep} contributed to Bengali VQA by developing a human-annotated Bengali VQA dataset and proposing a Top-Down Attention-based model. Collectively, these studies highlight the need for continued research in VQG and VQA for underrepresented languages, showcasing the potential of modern CNN architectures and pretrained models in advancing image-based question generation and answering.

\section{BanglaMemeEvidence Dataset Annotation}
\subsection{Meme Collection}
Our meme collection process involved a comprehensive search across a diverse range of sources, ensuring a rich and varied dataset. We scoured popular social media platforms such as Facebook, Instagram, and X (formerly known as Twitter) to capture the latest trends and viral content. Additionally, we delved into various online websites, forums, and communities known for their vibrant meme cultures. To ensure the dataset's breadth and relevance, we cast our net wide, encompassing memes spanning multiple thematic categories. We explored content related to Entertainment, where memes often riff on celebrity culture, movies, and television shows, providing a lighthearted and humorous take on popular media. In the realm of Politics, we curated memes that reflect the dynamic landscape of political discourse, capturing satirical commentary, political humor, and memes that encapsulate key events and figures in the political sphere. Our collection extended to Sports, where memes capture the passion, drama, and camaraderie of sporting events, as well as the lighter side of sports culture through memes celebrating iconic moments and poking fun at sporting rivalries. Education-themed memes offered insights into the world of academia, student life, and the challenges and quirks of learning environments, resonating with students and educators alike. In the realm of Technology, memes provided a humorous lens through which to explore the ever-evolving world of gadgets, software, and digital culture, offering witty observations on the quirks and frustrations of technology users.
\begin{figure*}[h]
\centering
\includegraphics[width=1.0\textwidth]{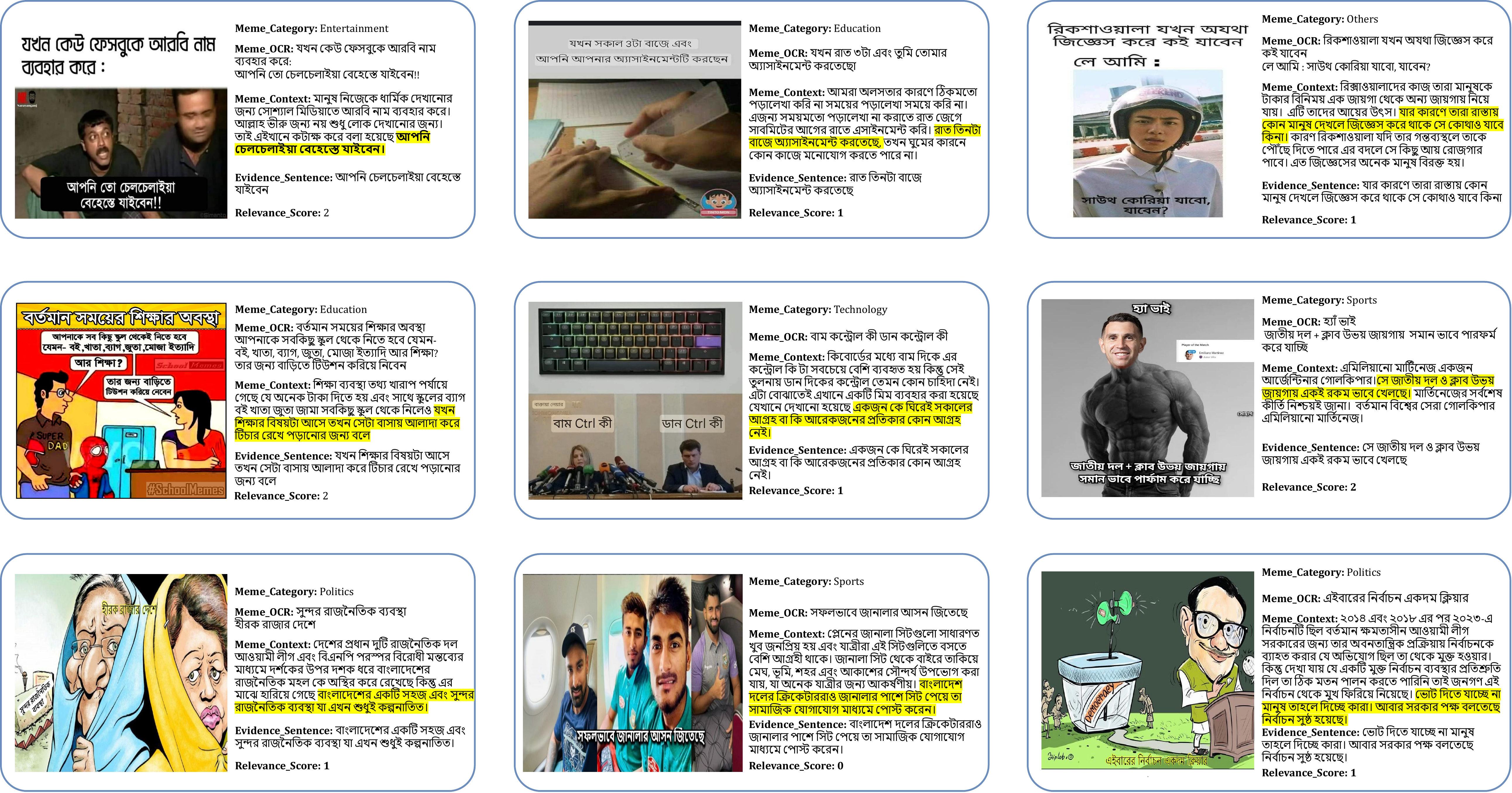}
\caption{Overview of BanglaMemeEvidence dataset across different meme categories}
\label{fig:ex}
\end{figure*}
\subsection{Context Document Curation}
In curating the contextual corpus corresponding to the memes collected, we employed a multifaceted approach. For some topics, we relied on Bangla Wikipedia \footnote{\url{https://bn.wikipedia.org/}} as a primary source, leveraging its wealth of information to provide evidence for the background of the memes. Additionally, we explored community-based discussion forums and question-answering websites such as Quora\footnote{\url{https://bn.quora.com/}}, as well as other general-purpose websites. Our thorough search spanned traditional media and digital publications, including newspapers\footnote{\url{https://www.prothomalo.com/}}, online article portals\footnote{\url{https://blog.muktomona.com/}}, and specialized media outlets\footnote{\url{https://ekattor.tv/}}, where we gathered in-depth analysis and expert commentary on meme-related events and phenomena. This comprehensive process ensured that each context document captured the essence of meme culture, offering valuable insights into its social, cultural, and political underpinnings.

\subsection{Dataset Description}
\begin{itemize}[label=$\circ$]
    \item \textbf{Meme\_ID:} Each meme is assigned a unique image ID for identification purposes.
    \item \textbf{Meme\_Category:} We have taken Politics, Sports, Entertainment, Education, Technology, and Others as meme categories. 
    \item \textbf{Meme\_OCR:} The Bengali text from the meme image is manually transcribed, ensuring accurate representation of the content.
    \item \textbf{Meme\_Context:} A concise description of the main Bengali context or theme depicted in the meme is provided.
    \item \textbf{Evidence\_Sentence:} This section includes a sentence or short text extracted from contextual documents, serving as evidence to support the interpretation or understanding of the meme.
    \item \textbf{Relevance\_Score:} Each evidence sentence in the dataset is tagged with a relevance score, indicating its relationship to the meme. This score operates on a scale where \textbf{(0)} denotes ``Not relevant,'' \textbf{(1)} signifies ``Partially relevant,'' and \textbf{(2)} represents ``Relevant.''
\end{itemize}

\begin{figure}[t]
  \includegraphics[width=1.0\textwidth]{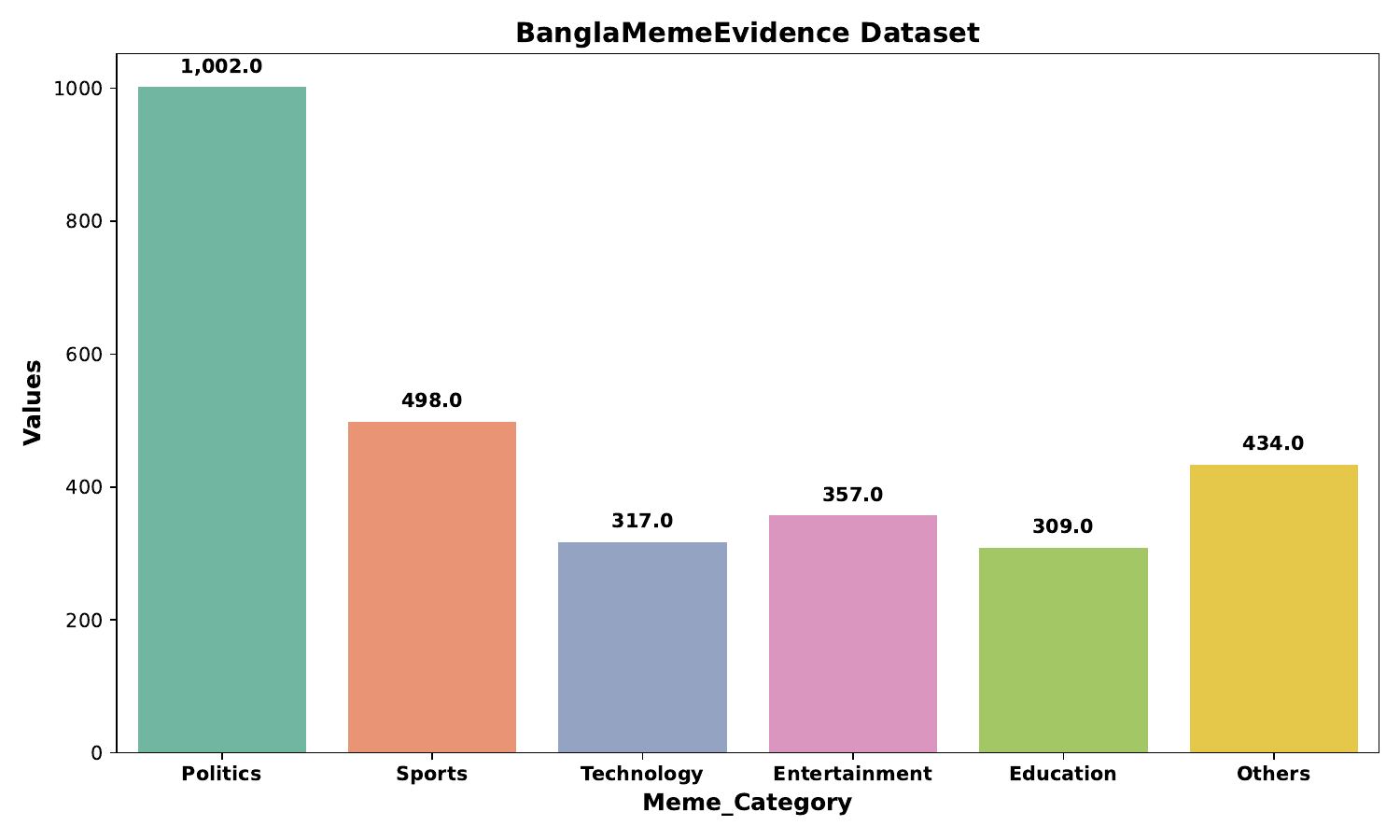}
  \caption{An illustration showcasing the diversity of meme categories, including Politics, Sports, Entertainment, and more from the BanglaMemeEvidence dataset.}
  \label{experiments}
\end{figure}

\subsection{Annotation Guideline}
We provided the following annotation guideline to the annotators:

\textbf{\textit{a) Comprehensive Comprehension:}} We ensure understanding of both the meme and its associated context before annotation. This comprehensive comprehension is crucial for accurate and insightful annotation.

\textbf{\textit{b) Semantics Alignment:}} We let the semantics of the meme guide the annotation process. By aligning with the intended meaning of the meme, our annotations maintain relevance and fidelity.

\textbf{\textit{c) Unit of Information:}} We recognize that self-contained, minimal units of information can serve as evidence. Each piece of evidence contributes to our understanding of the meme's background and significance.
    
\textbf{\textit{d) Non-Contiguous Evidence:}}  We acknowledge that valid evidence may not always occur contiguously. This understanding allows us to identify and annotate relevant information regardless of its spatial arrangement.
    
\textbf{\textit{e) Completeness Assurance:}} In cases where the context document does not support a meme, we diligently search for corroborating evidence from other established sources. This ensures the completeness and accuracy of our annotations.
    
\textbf{\textit{f) Caution with Ambiguity:}}   We exercise caution with ambiguous cases, opting to skip them to maintain the integrity of the annotation process. This approach prevents potential misinterpretations and ensures the reliability of our dataset.

\subsection{Annotation Process}
In our annotation process, we engaged the expertise of six male annotators, all hailing from Bangladesh and currently pursuing undergraduate studies. With ages ranging between 21 to 25 years, these individuals possessed not only a knack for meme creation but also an intricate understanding of meme culture, coupled with proficiency in navigating various social media platforms. Acknowledging the invaluable contribution of our annotators, we ensured fair compensation in accordance with Bangladeshi standards, recognizing the time and expertise they dedicated to the task. Equipped with a comprehensive set of guidelines, our annotators embarked on their mission: to scour context documents in search of succinct sentences that provided essential background information for each meme. These identified sentences, termed ``evidence sentences,'' served as the cornerstone of our dataset, offering profound insights into the genesis, significance, and cultural context of each meme. To ensure the reliability and consistency of the annotations, we implemented a rigorous process for addressing disagreements between annotators. Whenever annotators encountered differing interpretations of a meme or evidence sentences, they engaged in discussions to reach a consensus, guided by the annotation guidelines. In cases where disagreements persisted, a senior annotator, with deeper expertise in meme culture and contextual analysis, reviewed the conflicting annotations and made the final decision.
\begin{table}[h]
  \caption{Dataset Distribution Across Train, Test, and Validation Subsets Based on Relevance Score.} \label{stats}
  \setlength\tabcolsep{7pt}
  \centering
  \begin{tabular}{cccc}
    \hline
    \textbf{Relevance Score} & \textbf{Train} & \textbf{Test} & \textbf{Validation}  \\
    \hline
\centering    0  & 678 & 85 & 85 \\ 
  \centering  1  & 819 & 102 & 102 \\
   \centering 2  & 836 & 105 & 105\\
   \hline
  \end{tabular}
\end{table}

\subsection{Annotation Quality Maintenance}
In our annotation process, we utilized Fleiss Kappa \cite{fleshkappa} to ensure annotation quality. Fleiss Kappa assesses agreement among annotators, considering chance occurrences. It provides a score indicating agreement beyond chance. By monitoring Fleiss Kappa regularly, we maintained consistency and reliability in annotations. This practice ensured the dataset's accuracy and reliability for research and analysis. We achieved a Fleiss Kappa score of \textbf{0.87}, indicating strong agreement among annotators.

\subsection{Dataset Statistics}

We partitioned the dataset into three subsets: training (80\%), validation (10\%), and testing (10\%). This distribution ensures a balanced and representative split that supports effective model training, tuning, and evaluation. By reserving distinct portions of the data for each stage, we aim to promote generalization and minimize overfitting. Table \ref{stats} presents the data split by relevance score across train, test, and validation sets. Figure~\ref{fig:ex} illustrates representative examples from the dataset. The first meme critiques the superficial display of religious identity on social media through the adoption of Arabic names. The second highlights issues in the education system, where material provisions are present, but effective learning often relies on external tutoring. The third meme provides a humorous observation on technology usage, contrasting the frequent use of the left Ctrl key with the neglect of its counterpart on the right. The fourth meme celebrates the consistent athletic performance of footballer Emiliano Martinez at both the national and club levels. The fifth meme offers a pointed critique of electoral integrity, emphasizing low voter participation and the questionable fairness of the process. The sixth brings a lighthearted moment from sports, depicting Bangladeshi cricketers happily claiming window seats during air travel, capturing a relatable and simple joy. Figure~\ref{experiments} showcases its category-wise diversity.


\section{Implementation Details}
In our approach, \textit{\textbf{BengaliMemeEvidenceNet}}, for detecting explanatory evidence in memes, we explore multiple fusion techniques, including Early Fusion, Late Fusion, and Intermediate Fusion. By investigating and comparing the performance of these individual approaches, we develop a hybrid framework that combines the strengths of each fusion method. This exploration allows us to propose the best-performing fusion strategy, leveraging the advantages of each technique to achieve superior results in detecting explanatory evidence in memes.
\begin{figure*}
\centering
\includegraphics[width=1.0\textwidth]{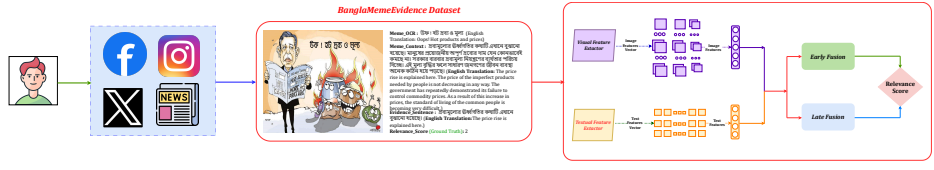}

\caption{Illustration of BengaliMemeEvidenceNet, a hybrid multimodal framework designed for identifying evidence in Bengali memes by leveraging related contexts and predicting relevance scores.
}\label{fig:workflow}
\end{figure*}
\textbf{Step 1) Text Preprocessing:} Before delving into analysis, we meticulously preprocess textual data extracted from various sources within memes, including \textit{Meme\_OCR}, \textit{Meme\_Context}, and \textit{Evidence\_Sentence}. This comprehensive preprocessing involves several steps aimed at ensuring the cleanliness and standardization of the text:

\begin{itemize}[label=$\circ$]
    \item \textbf{Punctuation Removal}: Removing punctuation marks such as exclamation points, question marks, underscores, and quotation marks. This can be mathematically represented as:
  \begin{equation}
    T_{\text{cleaned}} = \text{remove\_punctuation}(T)
 \end{equation}
    where \( T \) is the original text, and \( T_{\text{cleaned}} \) is the text with punctuation removed.

    \item \textbf{Whitespace Elimination}: Systematically eliminating any extraneous white spaces. This step is represented as:
  \begin{equation}
    T_{\text{cleaned}} = \text{remove\_whitespace}(T_{\text{cleaned}})
  \end{equation}
    where we remove any unnecessary whitespace from the text.

    \item \textbf{Emoji Removal}: Meticulously scanning and removing emojis to prevent noise and ambiguity. This is expressed as:
\begin{equation}
    T_{\text{cleaned}} = \text{remove\_emoji}(T_{\text{cleaned}})
 \end{equation}
    where emojis are identified and removed from the text.

    \item \textbf{Removing Non-Textual Content}: Remove any URLs, HTML tags, and special characters that are not relevant to the text analysis. This can be represented as:
\begin{equation}
    T_{\text{cleaned}} = \text{remove\_non\_textual}(T_{\text{cleaned}})
 \end{equation}
    where any non-relevant content is stripped from the text.

    \item \textbf{Spelling Correction}: Correct misspelled words to standardize the text. This is typically done by applying a spelling correction algorithm such as:
\begin{equation}
    T_{\text{corrected}} = \text{spell\_check}(T_{\text{cleaned}})
  \end{equation}
    where \( T_{\text{corrected}} \) is the text with corrected spelling.
\end{itemize}

\textbf{Step 2) Image Preprocessing:} To ensure uniformity and high quality across our dataset, we standardized the size of images extracted from memes to \( 224 \times 224 \) pixels, facilitating reliable and comparable analysis. Our image preprocessing pipeline includes several critical steps aimed at enhancing image quality and preparing the visual data for robust feature extraction and analysis:

\begin{itemize}[label=$\circ$]
    \item Normalizing the images to maintain consistency in pixel values. The normalization process is expressed as:
  \begin{equation}
    I_{\text{normalized}} = \frac{I - \mu}{\sigma}
  \end{equation}
    where \( I \) is the original image, \( \mu \) is the mean pixel value, and \( \sigma \) is the standard deviation of the pixel values.

    \item Applying edge detection algorithms (such as the Sobel operator) to highlight the boundaries within the images. This is represented as:
 \begin{equation}
    I_{\text{edges}} = \text{Sobel}(I)
  \end{equation}
    where \( I_{\text{edges}} \) contains the detected edges of the image.

    \item Applying noise reduction techniques, such as Gaussian smoothing, to reduce unwanted artifacts. This can be expressed as:
   \begin{equation}
    I_{\text{denoised}} = \text{Gaussian\_filter}(I)
    \end{equation}
    where \( I_{\text{denoised}} \) is the filtered image with reduced noise.

    \item Adjusting the contrast of the images to highlight significant features. This adjustment is expressed as:
\begin{equation}
    I_{\text{contrast\_adjusted}} = \text{adjust\_contrast}(I_{\text{denoised}})
  \end{equation}
    where \( I_{\text{contrast\_adjusted}} \) has improved contrast to better emphasize visual details.
\end{itemize}

\textbf{Step 3) Feature Extraction:} To capture the rich and intricate nuances of explanatory information within memes, we employ advanced pre-trained language models and cutting-edge image analysis techniques. Our feature extraction process is two-fold, focusing on both textual and visual data:
\begin{itemize}[label=$\circ$]
\item \textbf{\textit{Textual Feature Extraction:}} We utilize state-of-the-art (SOTA) pre-trained language models to extract semantic insights from various textual components of memes. The models we employ include mBERT \cite{mbert}, XLM-RoBERTa \cite{XLM}, and distilBERT \cite{Distil}. These models excel in understanding context and semantics, allowing us to accurately capture the subtleties and nuances inherent in the textual data of memes. We preferred mBERT, XLM-RoBERTa, and distilBERT due to their excellent multilingual capabilities, which are helpful when performing cross-lingual tasks and data collection from related languages. 
    
\item \textbf{\textit{Visual Feature Extraction:}} For the visual aspect, we utilize advanced models specifically tailored for meme analysis to extract relevant features from images. These models include Vision Transformers (ViTs) \cite{ViT}, Swin Transformer \cite{swing}, SwiftFormer \cite{swift}, PoolFormer \cite{pool}. Each model is trained to detect and emphasize significant visual elements that contribute to the explanatory content of memes.
\end{itemize}

\textbf{Step 4) Fusion Techniques:} After extracting text and image features, we employ fusion techniques tailored to each approach: Early Fusion \cite{EarlyFusion} and Late Fusion  \cite{LateFusion}.

\textbf{\textit{a) Early Fusion for Evidence Detection:}} In early fusion, we integrate features from both text and image modalities at a raw level and then feed them into a joint representation function. This joint representation is then used for further analysis. The equation for early fusion can be mathematically represented as:
\begin{equation}
\mathbf{Z}_{\text{joint}} = f_{\text{early}}(\mathbf{X}_{\text{text}}, \mathbf{X}_{\text{image}})
\end{equation}

Where:
\begin{itemize}[label=$\circ$]
    \item $\mathbf{Z}_{\text{joint}}$ represents the joint representation combining features from text and image.
    \item $f_{\text{early}}$ is the early fusion function.
    \item $\mathbf{X}_{\text{text}}$ denotes the feature representation extracted from text sources such as Meme\_OCR, Meme\_Context, and Evidence\_Sentence.
    \item $\mathbf{X}_{\text{image}}$ represents the feature representation extracted from the meme image.
\end{itemize}

\textbf{\textit{b) Late Fusion for Evidence Detection:}} In late fusion, predictions from text and image classification models are aggregated at a later stage to make the final decision. Each modality produces its prediction, and these predictions are combined using a fusion function. The equation for late fusion can be mathematically represented as:
\begin{equation}
\hat{y}_{\text{final}} = f_{\text{late}}(\hat{y}_{\text{text}}, \hat{y}_{\text{image}})
\end{equation}

Where:
\begin{itemize}[label=$\circ$]
    \item $\hat{y}_{\text{final}}$ represents the final prediction.
    \item $f_{\text{late}}$ is the late fusion function.
    \item $\hat{y}_{\text{text}}$ denotes the prediction obtained from the text classification model based on sources like Meme\_OCR, Meme\_Context, and Evidence\_Sentence.
    \item $\hat{y}_{\text{image}}$ represents the prediction obtained from the image classification model.
\end{itemize}

\textbf{\textit{c) Intermediate Fusion for Evidence Detection:}} In intermediate fusion, features from both text and image modalities are fused at an earlier stage, before the final classification decision is made. Rather than aggregating the predictions directly, the features from both modalities are combined and passed through a shared model to produce the final prediction. The fusion function in this case operates on the feature vectors of both modalities.

Let the feature vectors from the text and image models be denoted as \( \mathbf{f}_{\text{text}} \) and \( \mathbf{f}_{\text{image}} \), respectively. The intermediate fusion can be mathematically represented as:

\begin{equation}
\hat{y}_{\text{final}} = f_{\text{intermediate}}\left(\mathbf{f}_{\text{text}}, \mathbf{f}_{\text{image}}\right) = \mathbf{f}_{\text{text}} + \mathbf{f}_{\text{image}}
\end{equation}

Where:
\begin{itemize}[label=$\circ$]
    \item \( \hat{y}_{\text{final}} \) represents the final prediction after intermediate fusion.
    \item \( f_{\text{intermediate}} \) is the intermediate fusion function, which in this case is the element-wise sum of the feature vectors from both modalities.
    \item \( \mathbf{f}_{\text{text}} \) denotes the feature vector obtained from the text classification model, derived from sources such as Meme\_OCR, Meme\_Context, and Evidence\_Sentence.
    \item \( \mathbf{f}_{\text{image}} \) represents the feature vector obtained from the image classification model.
\end{itemize}

In this implementation, the intermediate fusion function \( f_{\text{intermediate}} \) combines the feature vectors from the text and image modalities by performing an element-wise sum of the two feature vectors.

\textbf{Step 5) Hyperparameter Tuning:}  Hyperparameter tuning plays a vital role in optimizing the performance of fusion strategies such as Early, Late, and Intermediate Fusion. For Early Fusion, the key hyperparameters include the learning rate ($\eta$), fusion weight ($w_{\text{fusion}}$), and dropout rate ($d$). The optimal learning rate $\eta_{\text{opt}}$ is determined by minimizing the loss function:
\begin{equation}
\eta_{\text{opt}} = \arg \min_{\eta} L(\theta; \eta)
\end{equation}
The fusion weight $w_{\text{fusion}}$ is tuned to balance the contribution of each modality in the fused feature vector, with the optimal value found as:
\begin{equation}
w_{\text{fusion}} = \arg \min_{w_{\text{fusion}}} L_{\text{fusion}}(X_{\text{text}}, X_{\text{image}}, w_{\text{fusion}})
\end{equation}
The dropout rate $d_{\text{opt}}$ is selected to prevent overfitting:
\begin{equation}
d_{\text{opt}} = \arg \min_{d} L(\theta; d)
\end{equation}
For Late Fusion, the model combines predictions from different modalities, and key hyperparameters include the learning rate ($\eta$), fusion weight ($w_{\text{fusion}}$), and batch size ($b$). The learning rate $\eta_{\text{opt}}$ is tuned similarly to Early Fusion, while the optimal fusion weight $w_{\text{fusion}}$ is selected by minimizing:
\begin{equation}
w_{\text{fusion}} = \arg \min_{w_{\text{fusion}}} L_{\text{fusion}}(y_{\text{text}}, y_{\text{image}}, w_{\text{fusion}})
\end{equation}
The batch size $b_{\text{opt}}$ is optimized to ensure efficient training:
\begin{equation}
b_{\text{opt}} = \arg \min_{b} L(\theta; b)
\end{equation}
For Intermediate Fusion, which combines features after extraction but before final decision-making, the learning rate ($\eta$), number of layers ($L$), fusion weight ($w_{\text{fusion}}$), and dropout rate ($d$) are crucial. The learning rate $\eta_{\text{opt}}$ and dropout rate $d_{\text{opt}}$ are optimized in the same way as in Early Fusion. The number of layers $L_{\text{opt}}$ is determined by minimizing the loss function:
\begin{equation}
L_{\text{opt}} = \arg \min_{L} L(\theta; L)
\end{equation}
The fusion weight $w_{\text{fusion}}$ is optimized similarly to Early Fusion:
\begin{equation}
w_{\text{fusion}} = \arg \min_{w_{\text{fusion}}} L_{\text{fusion}}(X_{\text{text}}, X_{\text{image}}, w_{\text{fusion}})
\end{equation}
Bayesian Optimization is employed to efficiently search for the optimal hyperparameters by modeling the loss function and predicting the best combination of hyperparameters. The overall optimization problem for all fusion methods can be formalized as:
\begin{equation}
\begin{aligned}
\theta_{\text{opt}}, \eta_{\text{opt}}, w_{\text{fusion}}, & \ d_{\text{opt}}, b_{\text{opt}}, L_{\text{opt}} \\
= \arg \min_{\theta, \eta, w, d, b, L} & \ L(\theta; \eta, w, d, b, L)
\end{aligned}
\end{equation}

where $L$ represents the loss function. This approach ensures that the optimal hyperparameters are selected to minimize the overall loss across different fusion strategies.

\textbf{Step 6) Evaluation of Meme Relevant Score Detection:} Evaluation involves assessing model performance in detecting explanatory evidence within memes using metrics such as accuracy, precision, recall, and F1-score. Precision measures the proportion of true positives among predicted positives, while recall evaluates the proportion of true positives among actual positives. The F1-score is crucial as it balances precision and recall into a single metric, which is particularly useful for handling imbalanced datasets where certain relevance scores are less frequent. This task includes classifying evidence into relevance scores: 0 for ``Not relevant,'' 1 for ``Partially relevant,'' and 2 for ``Relevant,'' making it challenging due to the imbalanced distribution of these scores.

\section{Result Analysis}

\begin{table}[ht]
   \caption{Performance Metrics of Various Fusion Approaches for Relevance Score Detection in Bengali Memes.}
   \centering
   \setlength\tabcolsep{5pt}
\begin{footnotesize}
  \begin{tabular}{p{2.0cm}lcccc}
    \hline
    \textbf{Approach} & \textbf{Models} & \textbf{Accuracy} & \textbf{Precision} & \textbf{Recall} & \textbf{F1-Score} \\
    \hline
     & ViT+mBERT & 0.67 & 0.65 & 0.66 & 0.66 \\
     & Swin Transformer+mBERT & 0.75 & 0.68 & 0.73 & 0.71 \\
     & SwiftFormer+mBERT & 0.74 & 0.65 & 0.73 & 0.69 \\
     & PoolFormer+mBERT & 0.72 & 0.75 & 0.70 & 0.72 \\
Early & ViT+XLM & 0.68 & 0.68 & 0.66 & 0.67 \\
Fusion & Swin+XLM & 0.74 & 0.72 & 0.67 & 0.69 \\
     & Swift+XLM & 0.74 & 0.65 & 0.72 & 0.69 \\
     & PoolFormer+XLM & 0.72 & 0.74 & 0.66 & 0.70 \\
     & ViT+DistilBERT & 0.66 & 0.71 & 0.66 & 0.68 \\
     & Swin+DistilBERT & 0.65 & 0.65 & 0.74 & 0.69 \\
     & Swift+DistilBERT & 0.75 & 0.72 & 0.67 & 0.70 \\
     & Pool+DistilBERT & 0.65 & 0.68 & 0.71 & 0.70 \\
\hline
     & ViT+mBERT & 0.68 & 0.72 & 0.65 & 0.68 \\
     & Swin Transformer+mBERT & 0.69 & 0.68 & 0.70 & 0.69 \\
     & SwiftFormer+mBERT & 0.72 & 0.73 & 0.71 & 0.72 \\
     & PoolFormer+mBERT & 0.70 & 0.71 & 0.69 & 0.70 \\
Late & ViT+XLM & 0.67 & 0.72 & 0.65 & 0.67 \\
Fusion & Swin Transformer+XLM & 0.71 & 0.70 & 0.72 & 0.71 \\
     & SwiftFormer+XLM & 0.68 & 0.69 & 0.67 & 0.68 \\
     & \textbf{BengaliMemeEvidenceNet} & \textbf{0.74} & \textbf{0.75} & \textbf{0.73} & \textbf{0.74} \\
     & ViT+DistilBERT & 0.73 & 0.74 & 0.72 & 0.73 \\
     & Swin Transformer+DistilBERT & 0.66 & 0.67 & 0.64 & 0.65 \\
     & SwiftFormer+DistilBERT & 0.71 & 0.69 & 0.73 & 0.71 \\
     & PoolFormer+DistilBERT & 0.65 & 0.64 & 0.66 & 0.65 \\
    \hline
  \end{tabular}
\end{footnotesize}
  \label{early-performance}
\end{table}
Table \ref{early-performance} present a comparative analysis of early and late fusion approaches for relevance score detection in Bengali memes, using F1-score as the primary evaluation metric. Among the early fusion models, Swin-m achieves the highest F1-score (0.71), while other models exhibit moderate performance. However, the late fusion method outperforms both strategies, with our proposed model, \textbf{BengaliMemeEvidenceNet }, achieving the highest F1-score of 0.74, surpassing all other architectures. This result underscores the effectiveness of BengaliMemeEvidenceNet  in capturing multimodal interactions, demonstrating its superiority in Bengali meme relevance detection over traditional fusion techniques. 

\section{Limitations}
Although our approach, BengaliMemeEvidenceNet, empirically outperforms several competitive baselines, we observe certain limitations in the modeling capacity towards BanglaMemeEvidence. As depicted in Table 6, there are three possible scenarios of ineffective detection: (a) no predictions, (b) partial match, and (c) incorrect predictions. The key challenges stem from the limitations in modeling the complex level of abstractions that a meme exhibits. These are primarily encountered in the following scenarios: 

\begin{itemize}[label=$\circ$]
    \item \textbf{Integration of Visual and Factual Knowledge:} A critical yet cryptic piece of information within memes often comes from the visuals, which typically require systematic integration of factual knowledge. BengaliMemeEvidenceNet currently lacks this capability, making it difficult to accurately interpret and explain the visual elements in memes.
    
    \item \textbf{Lexical Bias and Spurious Evidence:} The model is prone to picking up potentially spurious pieces of evidence due to lexical biasing within the related context. This can lead to incorrect predictions, as the model may focus on irrelevant or misleading textual features rather than the intended meaning of the meme.

    \item \textbf{Complex Abstractions:} Memes often exhibit a complex level of abstraction that is challenging for the model to capture. This includes subtle humor, cultural references, and context-specific nuances that require advanced reasoning capabilities beyond the current scope of BengaliMemeEvidenceNet.

    \item \textbf{Inadequate Contextual Understanding:} The model's current contextual understanding is often inadequate for fully grasping the meaning conveyed by memes. This limitation is particularly evident when memes rely on intricate social or cultural contexts that the model has not been trained to recognize.

\end{itemize}

\section{Future Works}

There are several areas where further research could significantly enhance the robustness and applicability of our models. Below, we outline future work that will build on our current findings:
\begin{itemize}[label=$\circ$]

\item \textbf{Dialect-Based Meme Annotation:} We will explore dialect-based meme annotation and subsequent performance analysis to account for variations in the Bengali language. This approach will improve model accuracy and effectiveness in diverse linguistic contexts, ensuring that regional dialects and linguistic variations are accurately represented and understood. 

\item \textbf{Banglish Meme Detection:} We aim to develop capabilities for detecting Banglish memes, which involve a mixture of Bengali and English, particularly prevalent in spoken communication by Bangladeshis. We will create methods to handle language switching within sentences or conversations, enhancing the model's ability to accurately interpret and explain Banglish memes. 

\item \textbf{Fine-Grained Semantic Role Analysis:} We will conduct a fine-grained analysis of the semantic roles in memes to explore nuances in connotation and interpretation. We will investigate how factors such as image composition, text placement, and meme format influence the perceived roles of entities like heroes, villains, and victims. This detailed analysis will provide deeper insights into how different elements of a meme contribute to its overall message. 

\item \textbf{Applications Beyond Memes:} We will investigate how the concept of visual semantic role labeling and natural language explanation generation can be applied to other forms of visual communication, such as advertisements, political cartoons, and social media posts. Expanding the application of these techniques will provide valuable tools for analyzing a wide range of visual media, enhancing our understanding of multimodal communication across different contexts.
\end{itemize}

\section{Conclusion}
We have introduced and explored the hybrid task of meme evidence detection, emphasizing the need for an in-depth understanding of the visual and textual semantics embedded in memes. By presenting BanglaMemeEvidence, a unique dataset of 2,917 annotated Bengali memes, we address a significant gap in meme analysis for low-resource languages. Our dataset includes rich annotations and relevance scores, providing a comprehensive resource for further studies in this domain. To effectively detect explanatory evidence within memes, we developed BengaliMemeEvidenceNet, a hybrid approach that integrates textual and visual features. This pioneering work not only sets a benchmark in the field of meme analysis for Bengali but also opens avenues for future research in other low-resource languages.

%
%
%
\bibliographystyle{unsrt}
\bibliography{custom}
%




\end{document}